\documentclass{article}

     \usepackage[nonatbib, preprint]{neurips_2020}

\usepackage[utf8]{inputenc} 
\usepackage[T1]{fontenc}    
\usepackage{hyperref}       
\usepackage{url}            
\usepackage{booktabs}       
\usepackage{amsfonts}       
\usepackage{nicefrac}       
\usepackage{microtype}      
\usepackage{amsmath}
\usepackage{graphicx}

\graphicspath{{Figs/}}

\title{Physically constrained short-term vehicle trajectory forecasting with naive semantic maps}

\author{
Albert Dulian\\
Dept. of Computer Science and Technology\\
The University of Hull\\
Kingston-upon-Hull, HU6 7RX\\
\texttt{A.Dulian-2013@hull.ac.uk}
\And
John C. Murray\\
Dept. of Computer Science and Technology\\
The University of Hull\\
Kingston-upon-Hull, HU6 7RX\\
\texttt{John.Murray@hull.ac.uk}
}

\begin{document}

\maketitle

\begin{abstract}
Urban environments manifest a high level of complexity, and therefore it is of vital importance for safety systems embedded within autonomous vehicles (AVs) to be able to accurately predict the short-term future motion of nearby agents. This problem can be further understood as generating a sequence of future coordinates for a given agent based on its past motion data e.g. position, velocity, acceleration etc, and whilst current approaches demonstrate plausible results they have a propensity to neglect a scene's physical constrains. In this paper we propose the model based on a combination of the CNN and LSTM encoder-decoder architecture that learns to extract a relevant road features from semantic maps as well as general motion of agents and uses this learned representation to predict their short-term future trajectories. We train and validate the model on the publicly available dataset that provides data from urban areas, allowing us to examine it in challenging and uncertain scenarios. We show that our model is not only capable of anticipating future motion whilst taking into consideration road boundaries, but can also effectively and precisely predict trajectories for a longer time horizon than initially trained for.

  
\end{abstract}

\section{Introduction}\label{sec:intro}
Human drivers present capabilities to infer other drivers' intentions from a number of observed sources e.g. visual cues, past actions, nearby traffic etc, and act accordingly to improve or maintain stable and safe on-road interactions \cite{underwood2003visual, martens2007familiarity}. For example, before making a lane change they examine if it is safe to execute a planned maneuver. Forecasting future trajectories however, is not a trivial task due to its highly non-linear and stochastic nature particularly as the environment complexity and prediction time-horizon increases. Nonetheless, achieving robust and accurate motion prediction is fundamental for improving a safety system's ability to effectively reason about numerous future states that are likely to occur. A variety of previously proposed approaches rely solely on using historical observations that describe vehicle motion to predict its future positions which leads to poor generalisation especially in urban environments. To overcome these limitations a model needs to also be able to understand the following:

\begin{enumerate}
    \item \textbf{On-road social interactions:} Predicted path might indeed be plausible and attainable by an agent, however, at the same time it might violate several social norms e.g. cutting into a other vehicle's lane.   
    \item \textbf{On-road physical constraints:} A model that is not capable of recognising road constraints might for instance predict a vehicle's future position outside of the road boundaries e.g. walkway.
    
\end{enumerate}

In this paper we address the latter issue and examine whether introducing spatial data that defines road topology can be utilised to develop a model that will present satisfactory capabilities on the task of short-term trajectory prediction in complex urban environments. We hypothesise that being able to further infer from spatial information will enable the model to, for instance, differentiate between drivable areas vs walkways and thereby model physical limitations and constrains of the road. 

Convolutional neural networks (CNNs) \cite{lecun1990handwritten} and Recurrent neural networks (RNNs) \cite{rumelhart1986learning} specifically long-short term memory (LSTMs) \cite{hochreiter1997long} have been extremely successful on tasks of image recognition \cite{zoph2018learning,krizhevsky2012imagenet} and sequence prediction \cite{graves2014towards,karpathy2014deep} respectively. We therefore propose a hybrid CNN-LSTM encoder-decoder architecture that first of all, learns spatial representation of the road from semantic maps, and secondly learns general motion of agents from observed data. The learned representation is then used to predict short-term motion of surrounding agents. We demonstrate the performance of our model on a publicly available dataset nuTonomy scenes (nuScenes) \cite{caesar2019nuscenes} and we further analyse predicted trajectories to understand which patterns cause most errors.

\
\section{Related work}\label{sec:related_work}
Forecasting future motion of nearby agents has been a topic of extensive study in recent years as the domain of autonomous transportation gained exponential interest both from numerous car manufacturers as well as various research institutions. Traditionally, a vehicle's future state has been recursively estimated with Kalman Filter \cite{kalman1960new} and noisy data from sensors. In addition, numerous techniques used Monte Carlo methods to generate a distribution of possible future trajectories of all agents in the scene and/or asses their risk \cite{broadhurst2005monte, eidehall2008statistical, althoff2011comparison}. Next, rather than strictly focusing on motion prediction, both \cite{lefevre2011exploiting} and \cite{firl2012predictive} tried to estimate the agent's future intended manoeuvres with the Bayesian Network and Hidden Markov Models (HMMs) respectively, and predict their future position using these estimations. An in-depth review of similar approaches is presented in \cite{lefevre2014survey}. 

However, a more recent body of work focused on using Deep Learning \cite{goodfellow2016deep} models for the task as previous approaches resulted in poor generalisation capabilities. These methods are often computationally expensive, thus not feasible for real-time prediction and are unable to accurately forecast a future location for longer time horizons e.g. 4-5 seconds. A number of proposed methods target this task from the perspective of forecasting future motion of humans, for instance Social LSTM \cite{alahi2016social} used LSTM-based architecture to model human interactions and their general movement in crowded spaces. Social GAN \cite{gupta2018social} builds on top of it by incorporating the generative model i.e. Generative Adversarial Network (GAN) \cite{goodfellow2014generative} to reason about human interactions and jointly predict multiple plausible future paths. SoPhie \cite{sadeghian2019sophie} extends it even further and combines attention mechanism \cite{xu2015show} within the model's architecture to yield socially as well as physically acceptable trajectories. 

Some of these ideas were later transferred to the domain of AVs', such as in \cite{deo2018convolutional} where an improved social pooling mechanism from \cite{alahi2016social} 
with a combination of manoeuvre classification was implemented to reason about neighbouring vehicles' movement and output distributions of future motions. Zhao \textit{et al.} \cite{zhao2019multi} also emphasised on the importance of modeling on-road interaction and proposed Multi-Agent Tensor Fusion (MATF) network that uses as its input observed past motion data of multiple agents and scene context, effectively allowing the decoder to output its predictions based both on encoded vehicle-to-vehicle interactions as well as physical constrains of the surrounding. The following approaches \cite{djuric2018uncertaintyaware, chou2019predicting, cui2019multimodal} present methodology that is marginally similar to ours i.e. scene context e.g. lanes, drivable areas, surrounding agents movement etc, was rasterised into top-down image and used as an input to CNN-based network to infer future positions of targeted agent. Chandra \textit{et al.} \cite{chandra2019robusttp} used segmentation algorithms \cite{redmon2016you, he2017mask} to identify and track agents in dense, heterogeneous traffic. Past trajectory data is then extracted from detected agents and passed into LSTM-CNN network that predicts future coordinates for up to 5 seconds. A similar approach was also presented in \cite{chandra2019traphic}, with an addition of modelling social interactions between various types of traffic agents.

\section{Data and methodology}\label{sec:data_meth}
\subsection{Problem formulation}
As commonly defined in the literature, vehicle trajectory prediction refers to the problem of predicting a sequence of agent's future movements given their prior observed data e.g. past positions, velocity etc. In this work we approach this issue by predicting the difference between agent's $x, y$ position at time $t$ and $t^{+1}$ denoted as a $\Delta x, \Delta y$ for $n$ seconds into the future, and then use the predicted change in position to compute the desired future motion. We argue that predicting $\Delta x, \Delta y$ rather than $x,y$ allows to:

\begin{enumerate}
    \item Generate data that closely follows a multivariate Gaussian distribution, thus enabling the model to learn simplified representation of dependable variables $\Delta x, \Delta y$
    \item Train the model that will present ability to generalise to unseen places. We believe that if the model was to directly predict $x,y$ coordinates it would fail if those would fall outside the learned distribution, therefore it is more beneficial to learn general motion and road patterns from observed data and semantic maps. 
\end{enumerate}

\subsection{Model's input/output} 
First of all, nuScenes provides 3D bounding box annotation data of agents within the scene, therefore we assume that the ego vehicle (a vehicle that executes prediction task) is equipped with a tracking algorithm capable of performing object detection. Next, for the currently tracked agent we define an input matrix:
\begin{equation}
    \mbox{\boldmath$X_{obs}$} = [\mbox{\boldmath$x_{obs}$}^{t-t\_obs}, \ldots, \mbox{\boldmath$x_{obs}$}^{t-1}, \mbox{\boldmath$x_{obs}$}^t]
    \label{eq:X_obs}
\end{equation}

representing past motion data observed for $t\_obs$ seconds where:
\begin{equation}
    \mbox{\boldmath$x_{obs}$}^t = [vel_x, vel_y, acc_x, acc_y, yaw]
    \label{eq:x_obs}
\end{equation}

is a vector of features at timestep $t$ containing velocity, acceleration and yaw angle respectively. In addition we define another input tensor that stores semantic map data as:
\begin{equation}
    \mbox{\boldmath$X_{spatial}$} = [\mbox{\boldmath$x_{spatial}$}^{t-t\_obs}, \ldots, \mbox{\boldmath$x_{spatial}$}^{t-1}, \mbox{\boldmath$x_{spatial}$}^t]
    \label{eq:X_spat}
\end{equation}

where $\mbox{\boldmath$x_{spatial}$}^t$ is a $256 \times 256$ RGB image representing an approximate 60 meter map chunk at time $t$. Each chunk is characterised by the following:
\begin{itemize}
    \item Semantic renderable road layers, specifically \textit{drivable\_area, road\_segment, lane and walkway} as defined by nuScenes.
    \item Past positions of the observed agent for $t\_obs$ seconds where position at time $t$ is rendered at the centre of the map
\end{itemize}

The model's output is then defined as a matrix:
\begin{equation}
    \mbox{\boldmath$\hat{Y}$} = [\mbox{\boldmath$\hat{y}$}^{t+1}, \mbox{\boldmath$\hat{y}$}^{t+2}, \ldots,  \mbox{\boldmath$\hat{y}$}^{t + t\_hor}]
    \label{eq:Y}
\end{equation}

containing predicted differences between positions for specified future time horizon $t\_hor$ where:
\begin{equation}
    \mbox{\boldmath$\hat{y}$}^{t+1} = [\Delta x, \Delta y]
    \label{eq:y}
\end{equation}
defines the predicted difference between spatial coordinates $x,y$ at time $t$ and $t^{+1}$. It is important to note that each subsequent $\Delta x, \Delta y$ pair does not define a difference between $x,y$ at time $t$ and $t^{+n}$ (i.e. it's overall change) but rather a difference between $x,y$ at time $t^n$ and $t^{n+1}$ (i.e. a change between timesteps), thus the $\Delta x, \Delta y$ always represents the difference of 1 timestep into the future. 
\newpage
\subsection{Model's architecture and representation}
\begin{figure}[ht]
    \centering
    \includegraphics[width=\linewidth]{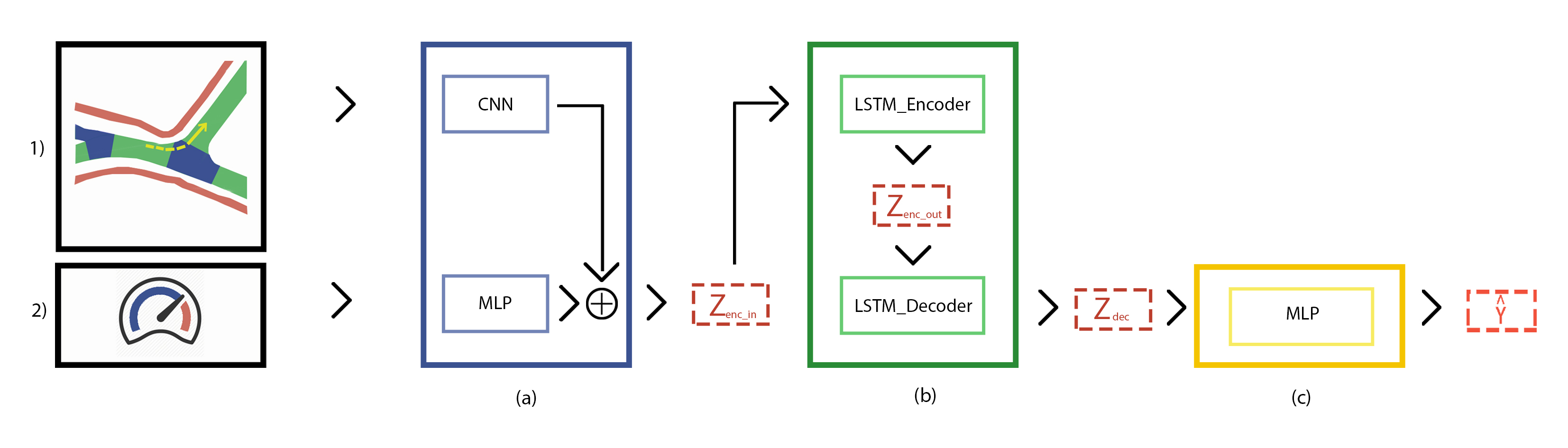}
    \caption{Summary of the network's architecture where 1) and 2) correspond to semantic maps and motion data input observed over specified time. The network is divided into three main blocks: (a) extracts spatial features from semantic maps and parameterises motion data, (b) temporal encoder-decoder, (c) predicts future motion from decoded data.}
    \label{fig:net_arch}
\end{figure}

Figure \ref{fig:net_arch} presents a simplified architecture of the proposed network, made up of three consecutive blocks. First, inspired by abilities of CNNs \cite{lecun1990handwritten} to extract salient features from spatial data we combine six convolutional operations with ReLU \cite{nair2010rectified} as our activation function in-between layers to obtain compressed, latent representation of road topology per $\mbox{\boldmath$x_{spatial}$} \in \mbox{\boldmath$X_{spatial}$}$. Furthermore, the agent's observed motion data $\mbox{\boldmath$x_{obs}$} \in \mbox{\boldmath$X_{obs}$}$ is parameterised with the first block's MLP \cite{goodfellow2016deep} module which also combines a single fully-connected layer with ReLU non-linearity. Note that throughout this paper the following convention \mbox{\boldmath$W_{*}$} is used to refer to weights matrices. The latent representation of \mbox{\boldmath$x_{spatial}$} as well as \mbox{\boldmath$x_{obs}$} is concatenated to form \mbox{\boldmath$z_{enc\_in}$} for $t\_obs$ timesteps resulting in a matrix \mbox{\boldmath$Z_{enc\_in}$}: 
\begin{equation}
    \mbox{\boldmath$z_{enc\_in}$} = concat(CNN(\mbox{\boldmath$x_{spatial}$}; \mbox{\boldmath$W_{spatial}$}), MLP(\mbox{\boldmath$x_{obs}$};\mbox{\boldmath$W_{obs}$})) 
    \label{eq:z_enc_in}
\end{equation}

\begin{equation}
    \mbox{\boldmath$Z_{enc\_in}$} = [\mbox{\boldmath$z_{enc\_in}$}^{t-t\_obs}, \ldots, \mbox{\boldmath$z_{enc\_in}$}^{t-1}, \mbox{\boldmath$z_{enc\_in}$}^t]
    \label{eq:Z_enc_in}
\end{equation}

Next, to overcome the issue of sequence generation we propose two additional blocks within the network. First, we define a single layer LSTM \cite{hochreiter1997long} encoder-decoder with shared weights and three LSTM cells where the computation mechanism of each cell is defined as:
\begin{equation}
    \begin{split}
        &\mbox{\boldmath$i$} = \sigma(\mbox{\boldmath$W_{ii}$}\mbox{\boldmath$x$} + \mbox{\boldmath$b_{ii}$} + \mbox{\boldmath$W_{hi}$}\mbox{\boldmath$h$} + \mbox{\boldmath$b_{hi}$}) \\   
        &\mbox{\boldmath$f$} = \sigma(\mbox{\boldmath$W_{if}$}\mbox{\boldmath$x$} + \mbox{\boldmath$b_{if}$} + \mbox{\boldmath$W_{hf}$}\mbox{\boldmath$h$} + \mbox{\boldmath$b_{hf}$}) \\ 
        &\mbox{\boldmath$g$} = tanh(\mbox{\boldmath$W_{ig}$}\mbox{\boldmath$x$} + \mbox{\boldmath$b_{ig}$} + \mbox{\boldmath$W_{hg}$}\mbox{\boldmath$h$} + \mbox{\boldmath$b_{hg}$}) \\ 
        &\mbox{\boldmath$o$} = \sigma(\mbox{\boldmath$W_{io}$}\mbox{\boldmath$x$} + \mbox{\boldmath$b_{io}$} + \mbox{\boldmath$W_{ho}$}\mbox{\boldmath$h$} + \mbox{\boldmath$b_{ho}$}) \\
        &\mbox{\boldmath$c^\prime$} = \mbox{\boldmath$f$} \odot \mbox{\boldmath$c$} + 
        \mbox{\boldmath$i$} \odot \mbox{\boldmath$g$} \\
        &\mbox{\boldmath$h^\prime$} = \mbox{\boldmath$o$} \odot tanh(\mbox{\boldmath$c^\prime$})
    \end{split}
    \label{eq:lstm_cell}
\end{equation}

where \mbox{\boldmath$i$}, \mbox{\boldmath$f$}, \mbox{\boldmath$g$}, \mbox{\boldmath$o$} are the gates of the cell i.e. input, forget, cell and output gates respectively, $\mbox{\boldmath$h$} \sim \mathcal{N}(0,1)$ is the initial hidden state and $\mbox{\boldmath$c$} \sim \mathcal{N}(0,1)$ defines the initial cell state, \mbox{\boldmath$x$} is the cell's input and its output is further defined as a tuple $(\mbox{\boldmath$h^\prime$}, \mbox{\boldmath$c^\prime$})$, $\sigma$ is the sigmoid activation function and finally, $\odot$ corresponds to Hadamard product. The forward pass of the LSTM encoder computes and updates its internal hidden states with \mbox{\boldmath$Z_{enc\_in}$} as follows:
\begin{equation}
    \mbox{\boldmath$Z_{enc\_out}$} = LSTM\_Encoder(\mbox{\boldmath$Z_{enc\_in}$; \mbox{\boldmath$W_{enc\_dec}$}})
    \label{eq:Z_enc_out}
\end{equation}

where \mbox{\boldmath$Z_{enc\_out}$} encapsulates the encoded representation of  \mbox{\boldmath$Z_{enc\_in}$} through time. Next, the final hidden state of LSTM encoder i.e. the last entry vector in matrix \mbox{\boldmath$Z_{enc\_out}$} which we refer to simply as \mbox{\boldmath$z_{dec\_hs}$} is recursively used as an input to LSTM decoder so that:
\begin{equation}
    \mbox{\boldmath$z_{dec\_hs}$}^{t+1} = LSTM\_Decoder(\mbox{\boldmath$z_{dec\_hs}$}^t; \mbox{\boldmath$W_{enc\_dec}$})
    \label{eq:z_dec}
\end{equation}

which is further used for $t\_hor$ timesteps to form:
\begin{equation}
    \mbox{\boldmath$Z_{dec}$} = [\mbox{\boldmath$z_{dec\_hs}$}^{t+1}, \mbox{\boldmath$z_{dec\_hs}$}^{t+2}, \ldots, \mbox{\boldmath$z_{dec\_hs}$}^{t+t\_hor}]
    \label{eq:Z_dec}
\end{equation}

where \mbox{\boldmath$Z_{dec}$} contains the decoded trajectory data. Lastly, we define the third block of our network as a two-layered MLP, yet again with ReLU activation between the first and second fully-connected layers. A single output of the MLP module is then defined as:
\begin{equation}
    \mbox{\boldmath$\hat{y}$} = MLP(\mbox{\boldmath$z_{dec\_hs}$}; \mbox{\boldmath$W_{out}$})
    \label{eq:mlp_out_y}
\end{equation}

where \mbox{\boldmath$\hat{y}$} denotes predicted future $(\Delta x, \Delta y)$, and therefore, the final output of the network for each $\mbox{\boldmath$z_{dec\_hs}$}\in\mbox{\boldmath$Z_{dec}$}$ is further denoted as: 
\begin{equation}
    \mbox{\boldmath$\hat{Y}$} = [\mbox{\boldmath$\hat{y}$}^{t+1}, \mbox{\boldmath$\hat{y}$}^{t+2}, \ldots,  \mbox{\boldmath$\hat{y}$}^{t + t\_hor}]
    \label{eq:mlp_out_Y}
\end{equation}

\subsection{Implementation and training details}

We use a $256 \times 256 \times 3$ as an input dimensions of the first block's CNN module and $5$ for the MLP module. The output of the last convolutional layer for defining the number of its feature maps is set to be $32$ with its latent space size reduced through convolution operations to $4 \times 4$. The MLP's input-output size remains unchanged as the aim is to parameterise observed motion data rather than reduce its dimensionality. Next, the input as well as the hidden size of the LSTM encoder-decoder block is equal to the size of concatenated output of both CNN and MLP modules i.e. $517$. Lastly, the MLP module within the third block has an input size set to $517$ and $258$ in its first and second fully-connected layers respectively. In addition, to improve generalisation capabilities of the network, we added the dropout layer to the LSTM decoder with a coefficient equal to $0.5$. The training as well as model's hyperparameters have been adjusted based on numerous empirical experiments and observations. Furthermore, we used $3\mathrm{e}{-4}$ as a learning rate with the Adam \cite{kingma2014adam} optimiser and Mean Square Error (MSE) loss function between true targets \mbox{\boldmath${Y}$} and the model's predicted output \mbox{\boldmath$\hat{Y}$}. The model was trained on a single Nvidia Titan XP GPU with PyTorch implementation. 

\section{Experiments and results}\label{sec:exp_results}
In this section we present results of our experiments carried out on the nuScenes dataset. This dataset contains large quantities of non-linear trajectories data extracted from real world driving scenarios and recorded in urban environments. Unfortunately, at the time of writing no other publicly available datasets offered the availability of semantic maps which is crucial for training purposes. To further observe the efficiency of our proposed model we compare its performance with the following baselines:

\begin{itemize}
    \item \textit{P-Net} - This variation of of the model is often used in the literature as a baseline. The network consists of LSTM layers that use observed motion data to predict the actual 2D coordinates for the defined time horizon.
    
    \item \textit{PD-Net} - An alternative to the above model with the same architecture, however, instead of directly predicting 2D coordinates, it predicts the position difference of positions between future timesteps.
    
    \item \textit{SM-Net} - This is a basic variant of our proposed network which takes as input past observed motion data as well as a single image representing semantic map at the last observed timestep.
    
    \item \textit{TSM-Net} - A final, improved version of \textit{SM-Net} with identical network architecture. However, instead of using single instance semantic map, this method combines both motion data and numerous semantic maps collected over specified past timesteps.  
    
\end{itemize}

In addition, it is crucial to mention that the input tensor \mbox{\boldmath$X_{obs}$} for both \textit{SM-Net} and \textit{TSM-Net} does not contain any data in regards to past positions of the agent as we believe that the model can learn and extract this from \mbox{\boldmath$X_{spatial}$}. However, for both \textit{P-Net} and \textit{PD-Net} the semantic map data is not provided, and we therefore believe that in order for these models to be able to fully infer from motion data it is necessary to modify \mbox{\boldmath$X_{obs}$} and include observed past positions.

Next, we randomly split the data (over 20k samples) into training, validation and test sets with $60\%/20\%/20\%$
ratio respectively to examine the model's capabilities on unseen trajectories. We report the prediction error of our model in meters with metrics commonly used in the literature for the specified problem i.e. \textit{Average Displacement Error} (ADE) and \textit{Final Displacement Error} (FDE). ADE is defined as an average $L2$ distance between $\mbox{\boldmath${y_{pos}}$} \in \mbox{\boldmath$\mathsf {Y_{pos}}$}$ and $\mbox{\boldmath$\hat{y}_{pos}$} \in \mbox{\boldmath$ \hat{\mathsf Y}_{pos}$}$ where \mbox{\boldmath$\mathsf Y_{pos}$} and \mbox{\boldmath$\hat{\mathsf Y}_{pos}$} denote 3D tensors containing all samples of actual and predicted positions for $t\_hor$ timesteps into the future respectively:

\begin{equation}
    ADE = \frac{1}{n}\sqrt{\sum_{i=1}^{n}( \mbox{\boldmath${y_{pos}}$}_i -  \mbox{\boldmath$\hat{y}_{pos}$}_i)^2}    
    \label{eq:ade}
\end{equation}

Next, in contrast to the former metric, the FDE is computed as an average $L2$ error of all samples at time $t\_hor$ i.e. final predicted position:

\begin{equation}
    FDE = \frac{1}{n}\sqrt{\sum_{i=1}^{n}( \mbox{\boldmath${y_{pos}}$}^{t\_hor}_i -  \mbox{\boldmath$\hat{y}_{pos}$}^{t\_hor}_i)^2}    
    \label{eq:fde}
\end{equation}

\subsection{Quantitative results}\label{sec:qt_res}
In this section we present results of our quantitative analysis starting with performance comparison of the previously defined models. By default we set $t\_obs = 8$ and $t\_hor = 8$ timesteps during training and inference (unless stated otherwise) which corresponds to $4$ seconds as the data is recorded at $2Hz$. Table \ref{tb:baseline_models} presents performance results of four different models measured with ADE and FDE metrics. 

As expected, the standard approach of directly predicting $(x, y)$ coordinates yields relatively high error for both metrics which we believe is caused by high variability and complexity of trajectories. Secondly, we observe a large improvement with use of \textit{PD-Net} on both ADE and FDE in comparison to \textit{P-Net} which further proves that our assumption of using $(\Delta x, \Delta y)$ over $(x, y)$ as the network's target values is valid. Next, results obtained from \textit{SM-Net} indicate that increasing the complexity of the baseline network and incorporating a single image of the semantic map produces slightly, however, not significantly better results. Lastly, we demonstrate almost $40\%$ decrease on ADE and FDE errors when comparing the performance of our final method \textit{TSM-Net} to \textit{SM-Net}. Results show the advantage of incorporating the semantic map data, and we therefore argue that the model has successfully learned the latent representation of road topology and agent's motion through time despite the stochastic nature of the environment.

\begin{table}[ht]
\caption{Performance of compared models measured with ADE and FDE in meters. All models predict trajectories for a fixed amount of time into the future i.e. 8 timesteps (4 sec) and are trained with 8 timesteps (4 sec) per observed data point.}
  \label{tb:baseline_models}
  \centering
  \begin{tabular}{lll}
    \toprule
    
    \textbf{Model}     & \textbf{Average Disp. Error}     & \textbf{Final Disp. Error}  \\
    \midrule
    P-Net              & 6.07               & 5.99  \\
    PD-Net             & 1.80               & 2.75  \\
    SM-Net             & 1.23               & 2.48  \\
    \textbf{TSM-Net}   & \textbf{0.77}      & \textbf{1.47}  \\
    \bottomrule
  \end{tabular}
\end{table}

Next, we re-train \textit{TSM-Net} by changing timesteps of observed data to $t\_obs=4$ and $t\_obs=6$ in order to compare model's performance against the default, proposed value $t\_obs=8$, note that throughout the training the prediction time horizon remains fixed i.e. $t\_hor=8$. In addition, during the inference mode we modify $t\_hor$ to numerous different settings to examine whether the model that is initially trained with $t\_hor=8$ can successfully predict $(\Delta x, \Delta y$) further into the future. The results are presented in Table \ref{tb:t_obs_hor}. 

First, as we anticipated, the last row with $t\_obs=8$ yields lowest errors. However, surprisingly, $t\_obs=4$ outperforms $t\_obs=6$ on all settings of $t\_hor$ except the initial $t\_hor=8$. In addition, both $t\_obs=4$ and $t\_obs=8$ perform slightly, although not significantly better when $t\_hor=10$, albeit being trained with $t\_hor=8$. Lastly, although both metrics result in growth of error as $t\_hor$ gets higher (excluding 2nd row) we can clearly see that the increase of the ADE with time is not considerably different until it reaches $t\_hor = 16$. Nevertheless, the FDE error changes dramatically which proves that predictions further into the future are more prone to error.

\begin{table}[ht]
  \caption{Performance of \textit{TSM-Net} on variety of $t\_obs$ and $t\_hor$ settings, values in each cell correspond to ADE/FDE errors. Note that each timestep setting is equivalent to $2 * sec$, e.g. $t\_hor=8$ corresponds to 4 seconds into the future. }
  \label{tb:t_obs_hor}
  \centering
  \begin{tabular}{llllll}
    \toprule
    
    \textbf{Observed Data} & \textbf{t\_hor = 8} & \textbf{t\_hor = 10} & \textbf{t\_hor = 12}  & \textbf{t\_hor = 14} & \textbf{t\_hor = 16} \\ 
    \midrule
    \textbf{t\_obs = 4}                         & 1.21/2.35      & 1.04/2.19       & 1.39/3.34        & 1.80/4.84       & 2.37/6.92  \\
    \textbf{t\_obs = 6}                         & 0.84/1.75      & 1.14/2.40       & 1.49/3.56        & 1.88/4.89       & 2.43/6.72  \\
    \textbf{t\_obs = 8}                & \textbf{0.77/1.47} & \textbf{0.65/1.38} & \textbf{0.86/2.25} & \textbf{1.17/3.51} & \textbf{1.67/5.34}  \\
    \bottomrule
  \end{tabular}
\end{table}

Lastly, Table \ref{tb:lit_models} presents a comparison of performance of our approach and several other methods from the literature. Firstly, it is crucial to emphasize that methods listed in Table \ref{tb:lit_models} were not directly re-trained on nuScenes dataset due to differences in requirements of input-output data that could not be satisfied. Therefore, we like to note that results presented in table below do not aim to provide direct comparison but rather act as a indication of performance. 

\begin{table}[ht]
\caption{Performance of our model against other models from the literature. Results presented below are taken from \cite{chandra2019traphic} where these models were trained for the task of vehicle trajectory prediction with a time horizon set to 5 seconds. We also use the variation of our model that yields a 5 seconds predictions for fair comparison. }
  \label{tb:lit_models}
  \centering
  \begin{tabular}{lll}
    \toprule
    \textbf{Model}     & \textbf{Average Disp. Error}     & \textbf{Final Disp. Error}  \\
    \midrule
    S-LSTM \cite{alahi2016social}       & 3.01               & 4.89  \\
    S-GAN  \cite{gupta2018social}       & 2.76               & 4.79  \\
    CS-LSTM \cite{deo2018convolutional} & 1.15               & 3.35  \\
    RobustTP \cite{chandra2019robusttp} & 0.96               & 1.97  \\
    TraPHic \cite{chandra2019traphic}   & 0.78               & 2.44  \\
    \textbf{TSM-Net}                    & \textbf{0.65}      & \textbf{1.38}  \\
    \bottomrule
  \end{tabular}
\end{table}

\subsection{Qualitative results}
\begin{figure}[ht]
    \centering
    \includegraphics[scale=.45]{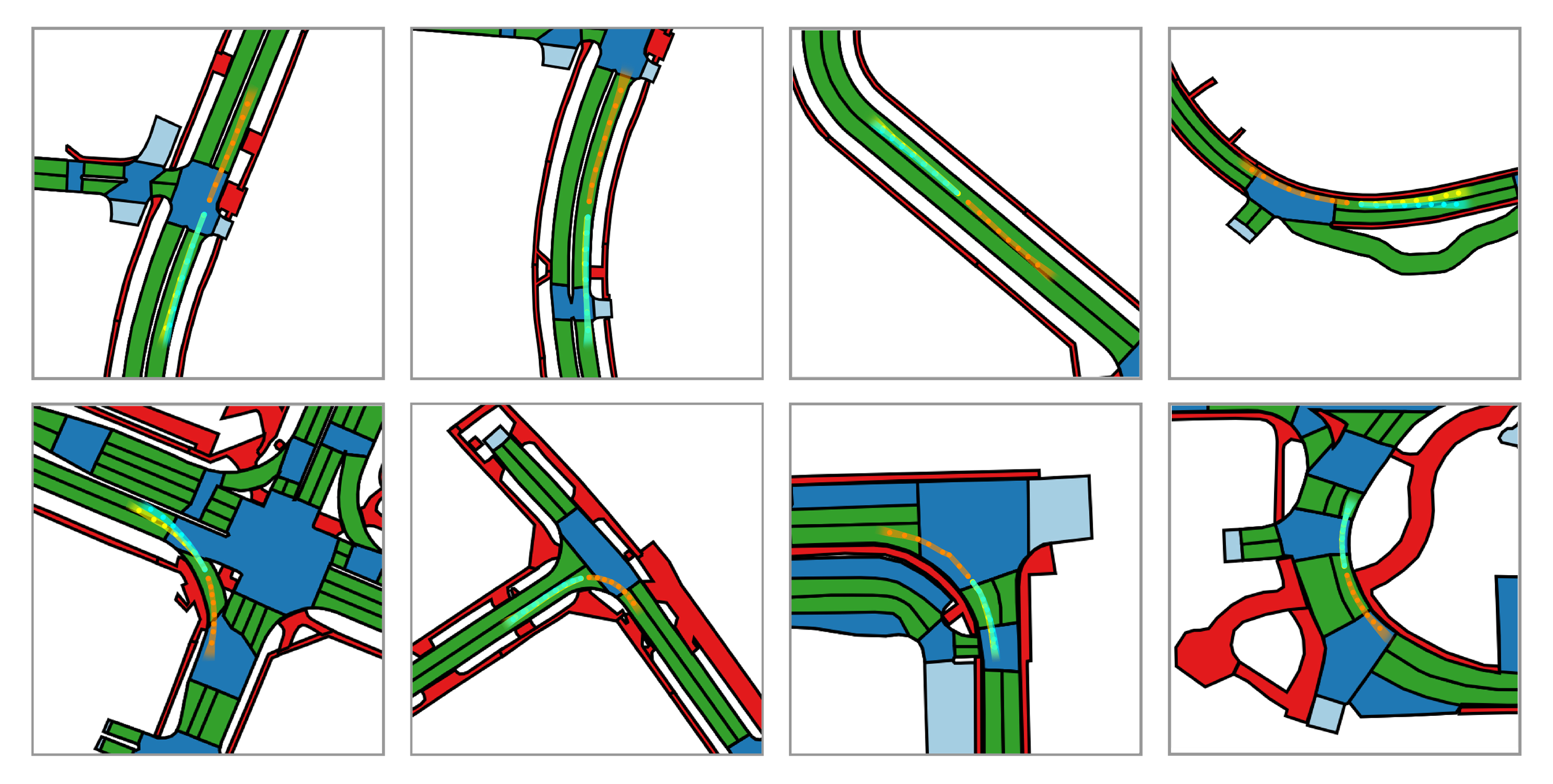}
    \caption{Exploration of predicted trajectories. The orange path represents observed route over time, whereas yellow and cyan corresponds to ground truth and model's prediction respectively. In few images the yellow paths are not fully visible, this is caused by the highly accurate prediction of the model which results with cyan route being overlaid on top of the ground truth.}
    \label{fig:qual_1}
\end{figure}

Results presented in section \ref{sec:qt_res} show that our proposed approach is capable of successfully predicting trajectories with a relatively low error even for a longer time horizon. To gain further insight into the model's abilities we visualise the agent's predicted motion directly on the semantic maps. Figure \ref{fig:qual_1} presents the first set of visualised trajectories. What becomes immediately apparent is the model's tendency to manifest a lower prediction error on linear motion, whereas if the future trajectory involves a slight bend or turn the deviation from ground truth tend to generally increase. 

\begin{figure}[ht]
    \centering
    \includegraphics[scale=.45]{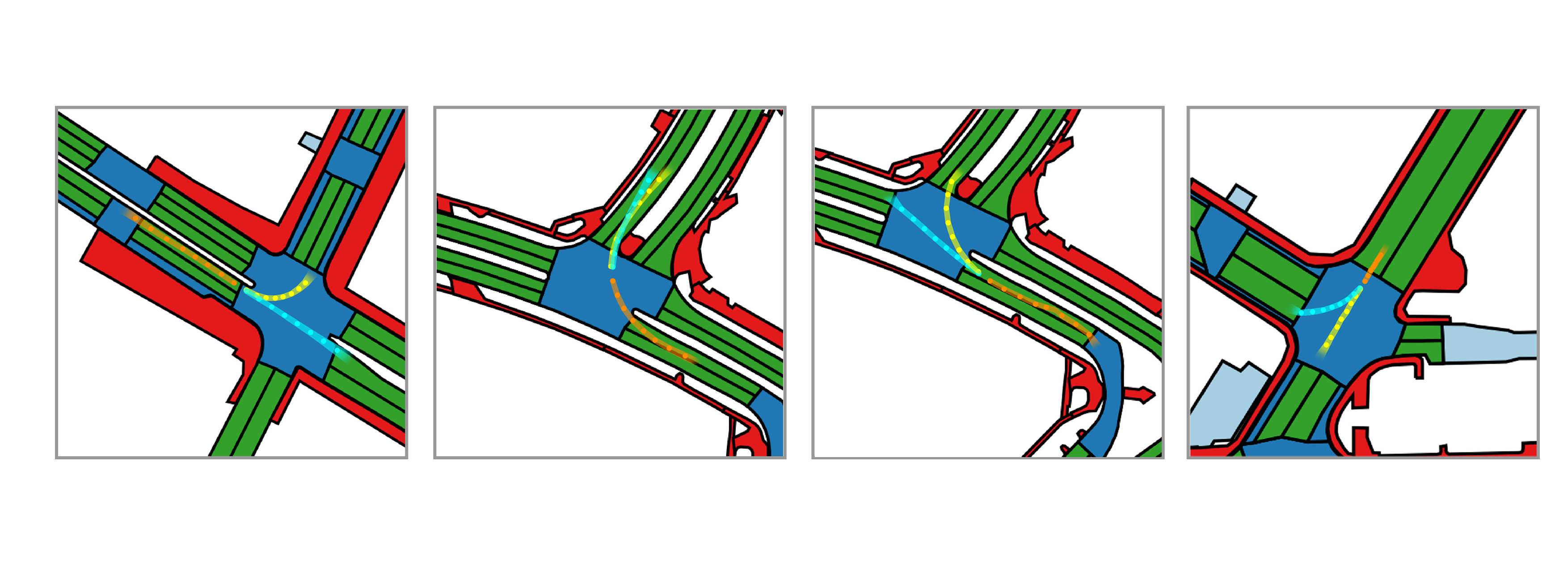}
    \caption{Further visualisation of model's output where anticipated trajectories greatly deviate from ground truth, albeit are still reasonable. Last two images demonstrate prediction where the error could have been highly influenced by training data that includes left and right hand traffic flow.}
    \label{fig:qual_2}
\end{figure}

Moreover, an interesting pattern emerged in several anticipated trajectories presented in Figure \ref{fig:qual_2}. The first image demonstrate a highly inaccurate prediction when compared to ground truth motion, nonetheless, the anticipated trajectory can indeed be classified as acceptable. Similarly, by looking at the second image it becomes apparent that although the route deviates from ground truth, it is still plausible and in reality could represent a lane change maneuver. Next, both third and fourth image resemble the same pattern as observed in the first prediction where a different path is taken, however, this time the model chooses a wrong side of the route. We strongly believe that this is caused by the fact that the data was collected in places where both left and right hand traffic occurs which the model is unable to distinguish.

\section{Conclusion}\label{sec:discussion}

In this work we present a novel approach towards the task of autonomous vehicle's trajectory prediction. Given an abstract, high level representation of the road topology as well as past motion data of the observed agent the proposed model has successfully managed to learn and extract its movement through time in complex, stochastic environment with relatively low error. Unlike prior techniques, our method is also able to recognise variety of different road boundaries and adjust generated path with respect to their physical limitations. Moreover, we discover that the model's ability to anticipate future motion is persistent even for longer time horizons despite not being explicitly trained to do so. Lastly, through further visualisation we also noticed that in number of situations predicted paths, although being incorrect in regard to the ground truth were in fact still plausible. 



\section{Broader impact}
In section \ref{sec:intro} we discussed that in order to create and develop a model that can produce highly accurate and safe results (especially in stochastic environments) with regards to the specified task it needs to not only understand the physical model of its surrounding but also numerous social norms (and a variety of other aspects that we have not included nor discussed) that traffic agents follow on a day-to-day basis. Thus, our work should not be considered as a singular, end-to-end method for motion forecasting, but rather an addition or an intermediate module that can be potentially deployed within safety systems to improve its performance. 

We also hope that a handful of ideas presented in this paper will shed a light on several factors regarding artificially intelligent based methods/systems (not only within intelligent transpiration domain) that need to be explored further. For instance, developing systems that can successfully act upon their environments in socially acceptable manner will hopefully also lead to an increase in trust between humans and machines.


\bibliographystyle{plain}
\bibliography{neurips_2020}

\end{document}